# E2MoCase: A Dataset for Emotional, Event and Moral Observations in News Articles on High-impact Legal Cases

**Candida M. Greco[1], Lorenzo Zangari[1], Davide Picca[2], Andrea Tagarelli[1]**
[1]University of Calabria, Rende, Italy
[2]University of Lausanne, Lausanne, Switzerland

## Abstract

The way media reports on legal cases can significantly shape public opinion, often embedding subtle biases that influence societal views on justice and morality. Analyzing these biases requires a holistic approach that captures the emotional tone, moral framing, and specific events within the narratives. In this work we introduce E2MoCase, a novel dataset designed to facilitate the integrated analysis of emotions, moral values, and events within legal narratives and media coverage. By leveraging advanced models for emotion detection, moral value identification, and event extraction, E2MoCase offers a multi-dimensional perspective on how legal cases are portrayed in news articles.

## 1 Introduction

The media plays a pivotal role in shaping public perception and discourse, especially concerning legal issues (Sadaf, 2011). Media narratives often present events imbued with emotions and moral undertones, influencing how individuals interpret and react to controversial topics. These narratives are composed of events that carry social or moral implications, described using emotionally charged language that triggers specific moral evaluations.

The analysis of human behavior requires a multifaceted approach that integrates different dimensions of values, which are "beliefs" that guide the interpretation of how individuals evaluate behaviors and events (Schwartz, 1992). Among these, morality stands as a fundamental aspect, critical for understanding behavioral responses, especially in the context of controversial and biased topics (Haidt and Joseph, 2004). However, relying solely on moral information can be limiting.

In fact, morality is often the driving force behind value-based judgments and emotional reactions (Raney, 2011). *Emotions* serve as immediate, intuitive responses that can precede rational deliberation, influencing how people interpret events. Moreover, *events* are actions that carry social or moral implications, and indeed they are described by words where the involved entities cause moral and emotional reactions; for example, witnessing an act of injustice can evoke emotions such as anger or empathy, which then shape the moral judgment and interpretation of the event.

The aspects of emotions, moralities and events, are deeply interconnected. Their study as concerns controversial arguments like media narratives on legal cases, can be valuable for analyzing human-driven values, since media narratives often frame events in emotionally charged ways, influencing moral evaluations and reinforcing or challenging existing value systems. Despite advancements in AI-based NLP tools like Large Language Models (LLMs), there is a lack of labeled data that jointly captures these three fundamental aspects: events, emotions, and moralities.

To face these challenges, we introduce a new semi-automatic dataset for **E**motional, **E**vent and **Mo**ral Observations in high-impact legal **Cases**, dubbed **E2MoCase**,[1] specifically designed to identify events, emotions and moralities in media narratives reporting on legal cases subject to bias.

Our developed dataset can be used for several NLP tasks based on one or more of its key-constituting dimensions (i.e., moralities, emotions, and events), but also to support related applications, such as bias detection, by training models on E2MoCase to recognize the moral, emotional, and factual characteristics of biased texts. This would be a step toward creating a more equitable landscape for legal and media activities, where the advancements in artificial intelligence can contribute positively to society.

We summarize our contributions as follows:

- We present E2MoCase, a novel dataset for analyzing emotions, moral values, and events

[1]We make our data available upon request.

in media coverage of legal cases.

- We show how the three aspects related to emotional tone, moral framing, and events arise from legal biased narratives.

- We provide an in-depth evaluation and validation of the proposed dataset through comparative experiments on morality and emotion detection.

The remainder of this paper is organized as follows: In Section 2, we review related work on emotion detection, morality analysis, and event extraction in legal and media contexts. Section 3 details the construction of the E2MoCase dataset, including the case selection, news retrieval, and annotation process. Section 4 provides an in-depth analysis of the dataset, exploring the distribution and correlations among emotions, moral values, and events. Section 5 provides an evaluation of the quality of the dataset's annotations.

## 2 Related Works

The current landscape of datasets tends to focus on either emotional analysis, moral reasoning, or event detection separately, with limited exploration of the intersection of these elements.

Some exploration into integrating these elements can be found in the works of Cherichi and Faiz (2016), who investigated the enhancement of social event detection using big data values, suggesting the potential for integrating various data types, including emotional and moral signals. Heron et al. (2018) provided a dataset specifically designed to capture moral emotions in dyadic interactions, offering a unique perspective on the interplay between moral considerations and emotional states. Gloor (2022) explored methodologies for measuring moral values from facial expressions, indicating a pathway towards the incorporation of these measurements into broader datasets that encompass event detection and moral evaluations. Zlatintsi et al. (2017) proposed COGNIMUSE, a dataset annotated with sensory and semantic saliency, events, cross-media semantics, and emotions, designed for training and evaluating algorithms for event detection, summarization, classification, and emotion tracking in videos. Moreover, the study by Hromic and Hayes (2014) based on constructing Twitter datasets that utilize various signals for event detection, provided insights into handling large-scale data where emotions and moral values are inherently intertwined.

In recent years, there has been a wide recognition that comprehensive data sets integrating moral, emotional and event dimensions are needed for effectively analyzing media narratives. As suggested in (Kepplinger et al., 2012), media depictions shape recipients' cognitions, emotions, and opinions, which extend beyond the original frames and are influenced by individual processing. In this regard, Zhang et al. (2024) introduced MORAL EVENTS, a dataset consisting of structured annotations of moral events across news articles from U.S. media reports, to study how moral values are implicitly conveyed through events rather than explicit moral language. Lei et al. (2024) proposed the EMONA dataset, containing news articles annotated with moral judgments of events to study how these opinions reflect ideological bias and media framing. Li et al. (2016) constructed a corpus of news articles for event-based emotion analysis where event anchor words are used as features for emotion classification.

The need for such multi-dimensional frameworks is even more exacerbated when it comes to media narratives that report legal cases. Such narratives are inherently multidimensional, involving not only the factual progression of events but also the emotional responses and moral judgments. Considering all these aspect is essential for training AI models capable of capturing the complexity of how legal issues are framed and perceived in public discourse. To the best of our knowledge, there is no dataset in the literature that answers to the combined need for capturing factual, emotional, and moral dimensions within legal media narratives. To address this gap in the literature, we introduce E2MoCase, a holistic dataset combining emotions, moral values, and events in media narratives reporting on high-impact legal cases. In capturing the complex interplay of emotions, moral values, and events in legal media narratives, E2MoCase provides an unprecedented, comprehensive resource for developing AI models that are sensitive to the multidimensional nature of public discourse on legal matters.

## 3 E2MoCase

E2MoCase is a novel dataset specifically designed to provide a multifaceted lens of legal narratives and press coverage through events, emotions and moral foundations. Figure 1 shows the construction process of E2MoCase, which consists of the following phases: **Case selection**, which was carried out semi-automatically to identify legal narratives that have attracted extensive media coverage; **News retrieval** from reputable newspapers concerning the cases selected in the case selection phase; **Dataset design and annotation**, whereby we automatically annotated news items at paragraph level with values of morality, emotions, and events, employing well-recognized domain-specific LLMs, namely *MoralBERT* (Preniqi et al., 2024), *EmoLLaMA* (Liu et al., 2024) and *GoLLIE* (Sainz et al., 2023).

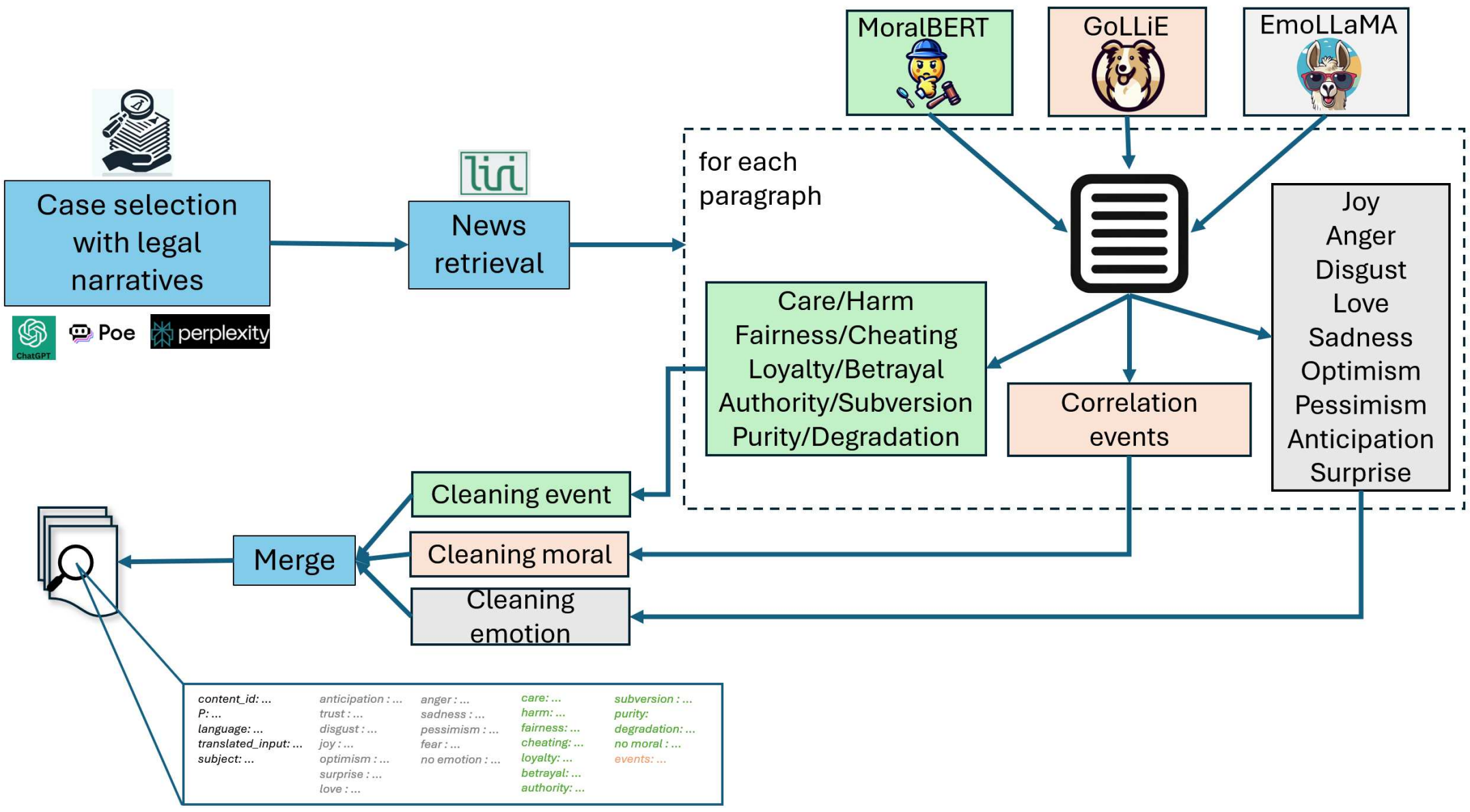

Figure 1: Workflow diagram illustrating the construction of E2MoCase.

Eventually, each data instance of E2MoCase includes (i) the subject of the case; (ii) the news article's identifier; (iii) the numerical identifier of the paragraph within the article; (iv) the paragraph translated into English (if the original language was different) and (v) its original language, (vi) the numerical scores associated with each of the emotion labels, (vii) the numerical scores associated with each of the morality labels, and (viii) a JSON string for each of the detected events.

In the following, we elaborate on each of the three phases performed for the construction of E2MoCase.

### 3.1 Case selection

We first selected more than 100 candidate cases related to legal matters that had significant media impact due to evidences of cultural biases, such as religious, political, gender, racial and media biases (Spinde et al., 2023). Specifically, we found such cases through both an extensive manual searching and the use of generative LLM tools with online search capabilities, particularly ChatGPT Plus,[2] Perplexity,[3] and Claude-3-Opus-200K.[4]

For each case, we manually verified its accuracy in terms of reported news, we ensured it had significant media impact and it was covered by reputable newspaper agencies. Example of prompts used to find the cases are shown in the **Appendix A**.

### 3.2 News retrieval

Given the candidate cases, we gathered the corresponding news data through the use of the Swissdox API,[5] which facilitates the retrieval of extensive volumes of Swiss media data for scholarly research. The Swissdox API provides news articles from the last 25 years, in 5 different languages: English, French, German, Italian and Romansh. It requires as main parameters: (i) the temporal range of the news to be gathered, (ii) the source news outlets, (iii) the language of the news, and (iv) a list of keywords that can be combined using the logical operators AND and OR. Considering parameter (i), we chose the year in which the case broke out, up to the year of this research study, i.e., 2024; for parameters (ii) and (iii), we selected any news outlets and language, while for (iv) we used the names of the main characters involved in the case combined with AND operator. Figure 5 in **Appendix B** shows an example of a query submitted to SwissDox.

[2] https://chatgpt.com/
[3] https://www.perplexity.ai/
[4] https://www.anthropic.com/claude
[5] https://liri.linguistik.uzh.ch/wiki/langtech/swissdox/start

Then, we analyzed all the candidate cases and selected a subset of 39 cases that we manually identified as crucial for encompassing a wide array of cultural biases (e.g., racial, gender). The 39 selected cases, along with their manually labeled detected biases, are detailed in **Appendix C**.

We finally built a corpus of 97,351 news paragraphs, which were translated into English to accommodate the predominance of English-based LLMs in recent research on moral, emotion, and event detection.

### 3.3 Dataset design and annotation

In our framework, we utilized a paragraph-level annotation to retain the contextual integrity of the original text. We believe that paragraphs represent a good balance between input length and semantic coherence preservation, retaining contextual integrity and supporting a fine-grained analysis. Specifically, each paragraph within E2MoCase was annotated with three golden labels, where each label consists of one or more values from each of the following types:

- **Emotional tone**, identifying the presence and intensity of specific emotions in the text: *joy*, *love*, *optimism*, *trust*, *sadness*, *disgust*, *pessimism*, *fear*, *anticipation*, *surprise*.
- **Moral trait**, which pertains to the underlying morality aspect according to the Moral Foundation Theory (Haidt and Joseph, 2004) considering their dyadic relationship: *care/harm*, *fairness/cheating*, *loyalty/betrayal*, *authority/-subversion*, *sanctity/degradation*.
- **Events occurrence**, which focuses on the events detected within the paragraph and the subjects involved. Events are defined as specific occurrences at a particular time and place involving one or more participants (Xiang and Wang, 2019), providing precise contextual information. In our setting, an event always includes a *mention*, i.e., a phrase/sentence describing the event. It may also include *entities*, i.e., the participants involved in the event, and their respective *roles*.

The above annotations were independently performed by using automated tools, as shown in Figure 1, and discussed in detail later in this section. We argue that the combined use of three types of labels to associate with news articles is crucial for a comprehensive understanding of the phenomena arising in the legal narratives carried out by modern media. Moralities, emotions, and events reflect the media's portrayal of a certain legal case.

**Emotion detection.** For detecting emotions in news paragraphs, we leveraged EmoLLMs (Liu et al., 2024), a series of open-source LLMs for comprehensive affective analysis. EmoLLMs are obtained by fine-tuning various LLMs (bart-large, T5-large, OPT-13B, Bloomz-7b1-mt, LLaMA2-7B, LLaMA2-chat-7B and LLaMA2-chat-13B) on the *Affective Analysis Instruction Dataset*, an instruction dataset based on the data from *SemEval 2018 Task 1: Affect in Tweets* (Mohammad et al., 2018). Following Mohammad et al. (2018), this dataset is constructed considering five prompt templates, each corresponding to a sentiment-based or emotion-based task, namely: (i) emotion intensity regression (EI-reg), (ii) ordinal classification of emotion intensity (EI-oc), (iii) sentiment regression (V-reg), (iv) ordinal classification of sentiment (V-oc), and (v) emotion classification (E-c). A comprehensive analysis of the performance of EmoLLMs on in-domain and out-domain data indicate that EmoLLMs perform exceptionally well in both regression tasks and classification tasks, surpassing the best model of SemEval 2018 leaderboard and achieving SOTA compared to the other open-source LLMs (Falcon, Vicuna, LLaMA-7B-chat, LLaMA-13B-chat).

Encouraged by these remarkable results, we opted to use the best-performing EmoLLM from the test set evaluations, i.e., EmoLLaMA-chat-13B, for the E-c task and the EI-reg task. This allowed us to identify a list of detected emotions from the input text (E-c), and to quantify the intensity of each detected emotion (EI-reg). In our experiments, we performed 8-bit quantization on the model, since it significantly reduces the computational requirements, allowing for faster inference times without substantial loss of accuracy.

**Moral classification.** We detected the moral values present in the news articles' paragraphs according to the Moral Foundation Theory (MFT) (Haidt and Joseph, 2004). The MFT framework identifies five key moral dimensions (*virtues*), each juxtaposed against its antithesis (*vices*): (i) *Care/Harm*, which emphasizes nurturing and kindness towards others; (ii) *Fairness/Cheating*, focused on recipro-cal altruism, celebrating virtues like honesty and integrity; (iii) *Loyalty/Betrayal*, highlighting the crucial aspect of group loyalty and the formation of strong relational bonds; (iv) *Authority/Subversion*, underscoring the importance of respecting and legitimizing authorities and traditions; (v) *Sanctity/Degradation*, which revolves around concerns for purity and the avoidance of moral or physical contamination. Our analysis seeks to identify the presence of one or more of these moral values for each paragraph, thus offering a structured approach to understanding the moral foundations of news articles. For this purpose, we utilized MoralBERT (Preniqi et al., 2024), a suite of BERT-based sequence classifiers, each fine-tuned for detecting and quantify moral values in social media data. In (Preniqi et al., 2024), both single-label and multi-label methodologies for moral predictions were explored, assessing both in-domain and out-domain moral inferences. Best results in out-of-domain contexts were obtained in the single-label scenario (i.e., one model designed for a specific moral value) and performing domain-adaptation with adversarial training. This is the configuration we have adopted for our analysis. To the best of our knowledge, MoralBERT represents the most valuable current approach for detecting morality.

**Event extraction.** We extracted events from each paragraph using GoLLIE (Sainz et al., 2023), an LLM derived from Code-LLaMa (Rozière et al., 2024) and fine-tuned to comply with annotation guidelines expressed as code-based instructions. By following the specific prompt described in GoLLIE (Sainz et al., 2023) for event extraction, we defined the target event encapsulating the guidelines and instructions as docstrings. Specifically, we focused on what we called as *Correlation Event*, i.e., any dynamics, state change or relationship involving one or more entities. The definition of the Correlation Event and some examples of the extracted events are shown in **Appendix D**. In our experiments we used Code-LLama with 13B parameters and 8 bit quantization. Note that we also tested InstructUIE (Wang et al., 2023) for the task of event extraction, which is a fine-tuned version of Flan-T5 with 11B parameters. However, we empirically found out that the results provided by GoLLIE were more accurate and consistent across different paragraphs, i.e., same output format. Note that, in the following, we will use the term "event" to intend correlation events.

| | E2MoCase | E2MoCase_noEvents | E2MoCase_full |
|---|---|---|---|
| # paragraphs | $50,975$ | $46,276$ | $97,251$ |
| avg # tokens | $275.106 \pm 245.303$ | $139.402 \pm 220.950$ | $210.532 \pm 243.647$ |
| avg # emotions | $1.164 \pm 0.757$ | $1.634 \pm 0.680$ | $1.678 \pm 0.657$ |
| avg # morals | $3.517 \pm 3.870$ | $1.773 \pm 1.644$ | $2.795 \pm 2.424$ |
| avg # events | $3.597 \pm 2.940$ | $0.0 \pm 0.0$ | $1.885 \pm 2.785$ |

Table 1: Statistics of E2MoCase and its variants: E2MoCase_noEvents, which is obtained by removing paragraphs that do not contain events, and E2MoCase_full, which also includes paragraphs that do not contain events.

## 4 Analysis of E2MoCase

News articles collected from SwissDox come from web pages, which may include various events of the day or insertions between paragraphs that are unrelated to the case of interest. To address this issue, we have identified these outliers by leveraging emotional information, and manually filtered out them. The full approach is described in **Appendix E**.

Table 1 shows the statistics of E2MoCase with its variants, i.e., E2MoCase containing the paragraphs with no detected events (E2MoCase_noEvents) and E2MoCase with all the paragraphs (E2MoCase_full). We can notice that the event extraction process identified events in only 50,975 paragraphs, resulting in the exclusion of nearly half of the initial paragraphs. We can also observe high variability in the paragraph length in terms of number of tokens (counted using WordPiece tokenizer (Devlin et al., 2018)) in all the variants; in addition, the average number of detected emotions per paragraph is 1.16 in all cases, whereas the average number of moral values is significantly lower in paragraphs without events compared to those with events. This also indicates that the presence of events, involving actions, decisions or situations, causes a stronger moral response.

We further analyze E2MoCase from the perspective of each type of label to gain insights about the underlying media narratives. Figure 2a and 2b show the distribution of emotions and morality traits, respectively. On the one hand, it can be observed that the most frequently detected emotions are *anger* and *disgust*, which are also strongly correlated (see Figure 10 in **Appendix F**). On the other hand, the distribution of moral values is much more balanced, with a predominance of *cheating*, *subversion*, and *betrayal* values, which are also strongly correlated (see Figure 9 in **Appendix F**). From Figure 3, we observe that *cheating*, *betrayal*, *subversion* and *degradation* moral values show a significant positive correlation with emotions such as *anger* and *disgust*. This suggests that narratives involving negative moral values are likely to evoke strong negative emotions and vice versa. Additionally, when considering moral values like *purity*, *fairness* and *care*, we also observe a slight positive correlation with negative emotions such as *anger* and *disgust*. This counter-intuitive finding may indicate that media narrative arousing these positive moral values are perceived as particularly sensational, thereby eliciting strong negative emotional responses. Nonetheless, as depicted in Figure 4, the events behind the news selected for E2MoCase, are quite controversial such as *murder*, *shoot*, *rape*. This is expected given the topics covered by the collected news articles. Additionally, we observed that in the paragraphs lacking any emotions or moral dimensions, which are approximately 3% of the total, the associated events include actions like *buy*, *sign*, *submit*, *announce*, *move*, *win*, and *increase* (see Figure 12 in Appendix F). It is likely that these events do not evoke strong moral or emotional responses because they are often perceived as routine or neutral actions, devoid of inherent effective implications.

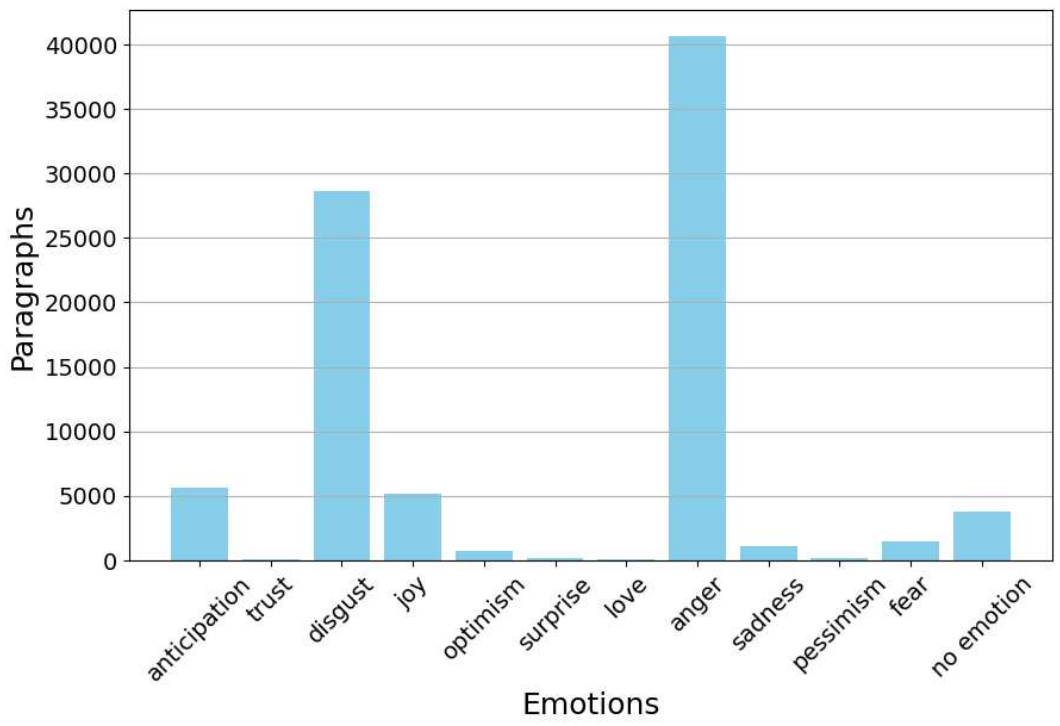

(a)

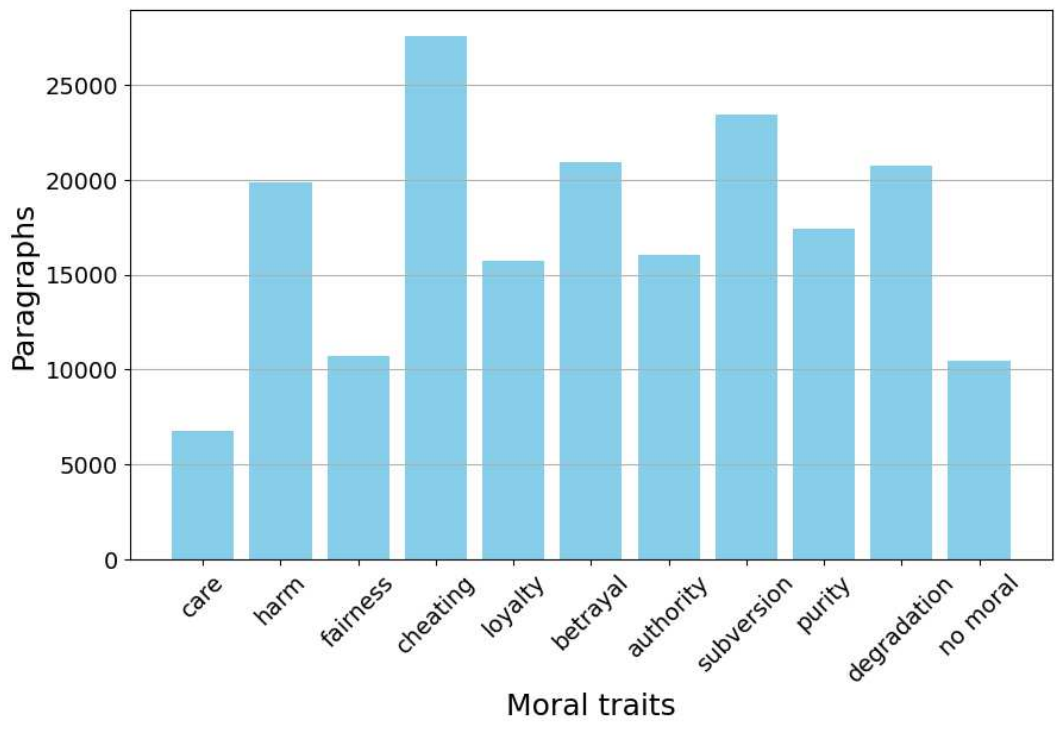

(b)

Figure 2: (a) Distribution of emotions and (b) Distribution of moral values

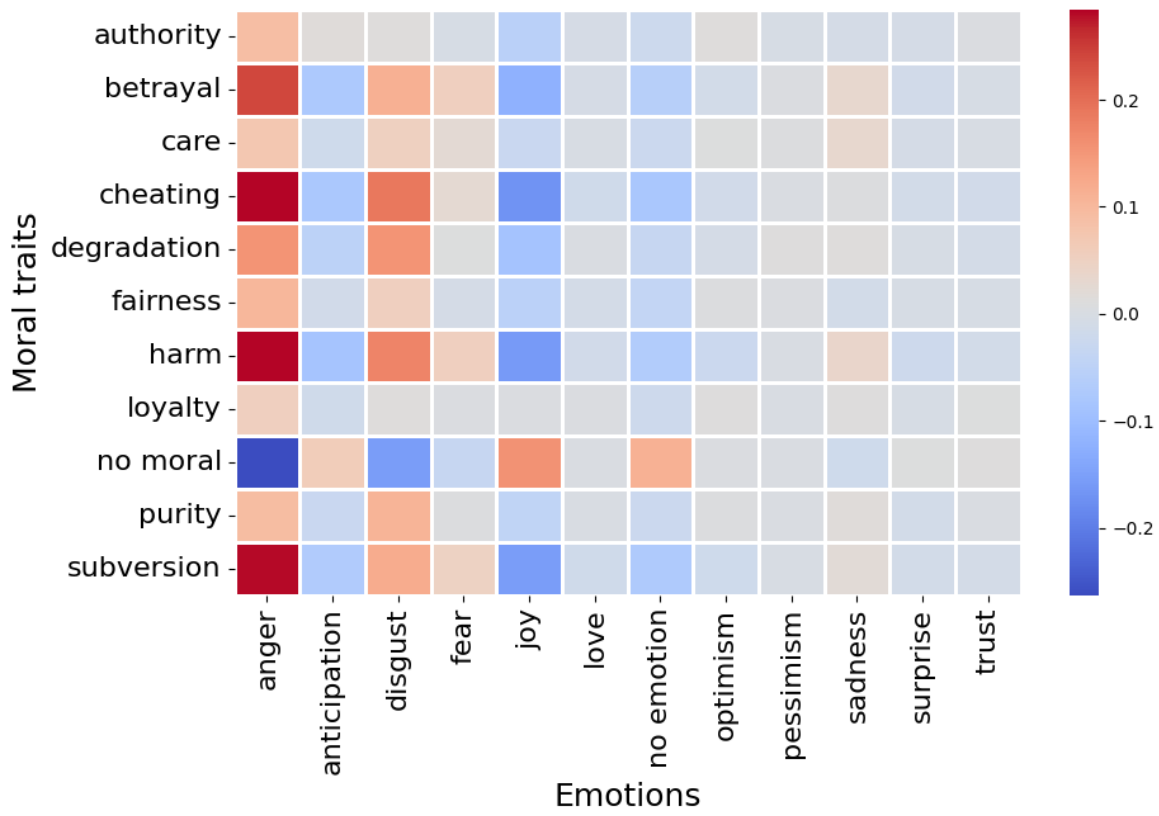

Figure 3: Pearson's correlation between emotions and moral values.

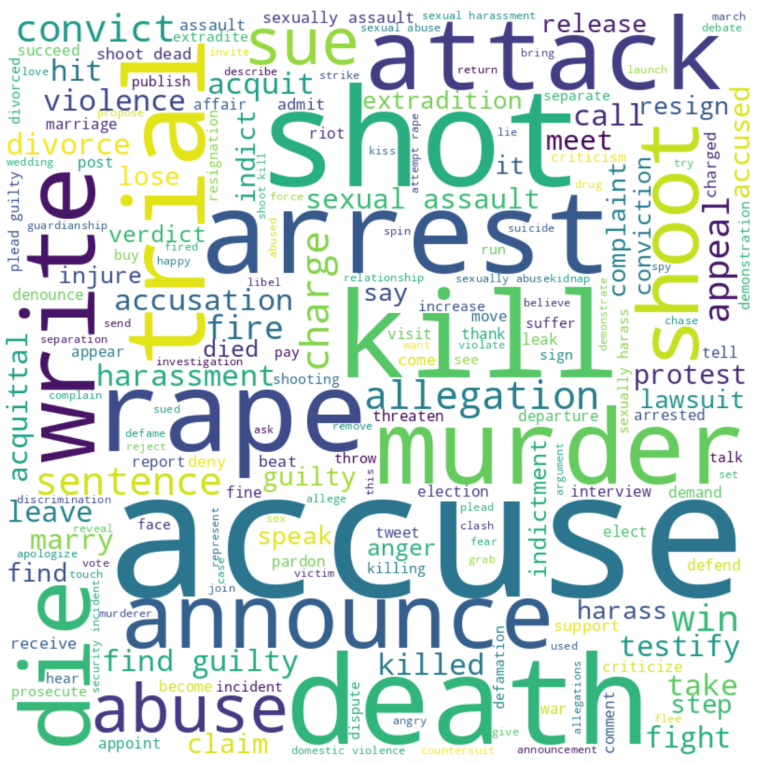

Figure 4: Word cloud of the detected events.

**Summary.** The analysis of E2MoCase reveals that the event extraction process identified events in 50,975 paragraphs, excluding nearly half of the initial paragraphs (i.e., E2MoCase_full). Paragraphs show high variability in length, with an average of 1.16 detected emotions and an average number of moral values significantly lower in paragraphs without events compared to those with events. There is an imbalance in the distribution of emotions, with negative emotions being more prevalent, while moral values are more evenly distributed. Negative emotions, particularly *anger* and *disgust*, frequently co-occur and are strongly associated with negative moral values such as *cheating*, *betrayal*,*subversion*, and *degradation*. Additionally, events depicted in E2MoCase, like *murder* and *rape*, align with the dataset's focus on controversial legal cases, explaining the prevalence of intense moral and emotional responses.

## 5 Validation of E2MoCase

Since the emotions, moralities, and events in E2MoCase have been obtained automatically (i.e.,

through MoralBERT, EmoLLaMA, GoLLiE), it is essential to provide some guarantee of the reliability of these labels. To this purpose, our approach was to train models on E2MoCase for tasks of moral value detection and emotion detection in order to validate the test performances on each of the three dimensions separately.

**Validation of moral labels.** Considering the moral dimensions, we examined the validity of the labels in E2MoCase for detecting moral values in human-annotated texts. Specifically, we fine-tuned a BERT base model (`bert-base-uncased`) on E2MoCase for morality classification and assessed the model's performance on two state-of-the-art datasets for moral values classification, based on the Moral Foundation Theory:

- *Moral Foundations Reddit Corpus* (MFRC) (Trager et al., 2022): a collection of 16,123 manually annotated Reddit comments, extracted from morally-relevant subreddits about US and French politics, and everyday moral life. Note that the morality annotations in MFRC is intended without the virtue-vice separation (e.g., Care/Harm is a unique label).
- *Moral Foundations Twitter Corpus* (MFTC) (Hoover et al., 2020): a corpus of 35,108 manually annotated tweets from several domains discussing morally-relevant and popular topics, such as Black Lives Matter, Hurricane Sandy, MeToo movement.

We considered the task of moral value classification in a *multi-label setting*. We binarized the MoralBERT-based numerical scores associated with morality values in E2MoCase (i.e., scores equal to or greater than 0.5 were mapped to 1, otherwise to 0). We performed a K-Fold cross-validation strategy, with $K = 5$. Following other works (Trager et al., 2022; Araque et al., 2020), for the MFTC dataset we adopted the common approach of reducing the moral labels to 5, without distinguishing between virtues and vices. The fine-tuning was conducted for 5 epochs with a batch size of 8, using the Adam optimizer, and a learning rate of 5e-5.

In Table 2, we compare the results on MFRC and MFTC test sets obtained by BERT trained on E2MoCase (dubbed BERT$_{E2Mo}$), BERT trained on MFTC (BERT$_{MFTC}$) and BERT trained MFRC (BERT$_{MFRC}$), under the same settings described above. We report the mean F1 score for each moral label, along with an overall evaluation of morality detection using the averaged micro F1 score (overall F1). Bold values, resp. underlined values, correspond to the best. resp. the second-best, results for each moral value (analogous remarks holds for the subsequent tables).

As expected, the *in-domain* models, i.e. models trained and tested on the same dataset, generally achieve the best performance. More interestingly, in most cases BERT$_{E2Mo}$ outperforms its *out-of-domain* counterparts (i.e., BERT$_{MTFC}$ for MFRC and BERT$_{MFRC}$ for MFTC), which are trained on human-annotated labels. In few cases, BERT$_{E2Mo}$ even outperforms the in-domain models (e.g., for the Loyalty label in the MFRC dataset).

Based on the above results, there is a basis for asserting that the morality labels of E2MoCase are consistent with human judgment, given that the model trained on these labels performs sufficiently well in detecting human-provided labels.

In addition, Figure 13 in the Appendix F provides further evidence on the validity of the moral labels of E2MoCase, showing that: (i) texts with the same label are sufficiently similar to each other and distinct from texts with different labels, and (ii) the labeling patterns across E2MoCase, MFRC, and MFTC datasets are consistent to each other.

**Validation of emotion labels.** Similarly, we tested the validity of the emotion labels of E2MoCase. In this case, we used the data from (Mohammad et al., 2018) as a comparison, for two main reasons: the emotion labels in this dataset are human-annotated and correspond to those of E2MoCase. As described in Section 3.3, the aim of Mohammad et al. (2018) is to evaluate various computational methods for detecting affective states, such as emotions and sentiments, expressed in tweets. We employ the data designed for emotion classification (E-c) task, hereinafter referred to as *SemEval*, containing about 11k instances. However, looking at Figure 2a, we observe that the labels *trust*, *surprise*, *love* and *pessimism* are almost absent in E2MoCase. Since it would not be possible to validate E2MoCase on these labels, we excluded them from the validation process and fine-tuned a BERT-base-uncased model on both E2MoCase and *SemEval* for the remaining labels (*anger*, *disgust*, *fear*, *joy*, *optimism*, *sadness*). We employed the same training settings as previously described for morality classification. However, here the emo-tion labels of E2MoCase were binarized by selecting any emotion score greater than 0. This is due to the fact that, as shown in Figures 2a and 2b, while the morality scores in E2MoCase are evenly distributed, the emotion scores in E2MoCase are highly imbalanced and the values are mostly below 0.5 (see Figure 11 in the **Appendix F**). Firstly, we tested the performance of BERT trained on E2MoCase, referred to as BERT$_{E2Mo}$, and BERT trained on SemEval, referred to as BERT$_{SemEval}$, on the SemEval test set to compare in-domain and out-of-domain performance. Additionally, we evaluated the performance of both models on other three emotion classification datasets, namely:

- *DailyDialogue* (Li et al., 2017): a manually-labeled dataset of conversations about daily life, which consists of a test set of about 1k instances with seven emotion labels (*anger*, *disgust*, *fear*, *joy*, *sadness*, *surprise* and *no-emotion*).

- *GoEmotion* (Demszky et al., 2020): a dataset consisting of Reddit comments labeled with 27 human-annotated emotions. From the test set, we selected only those instances labeled with the emotions that correspond to the non-underrepresented labels in E2MoCase, for a total of 880 instances.

- *AffectiveText* (Strapparava and Mihalcea, 2007): a dataset containing news headlines from sources like Google News, CNN, and newspapers, with a development set of 250 annotated headlines. It includes six human-annotated emotion categories: *anger*, *disgust*, *fear*, *joy*, *sadness*, and *surprise*.

| | MFRC | | | | | | MFTC | | | | | |
|---|---|---|---|---|---|---|---|---|---|---|---|---|
| | *Care* | *Fairness* | *Loyalty* | *Authority* | *Purity* | overall F1 | *Care* | *Fairness* | *Loyalty* | *Authority* | *Purity* | overall F1 |
| BERT$_{MFTC}$ | 0.407 | 0.358 | 0.189 | 0.259 | 0.199 | 0.282 | **0.770** | **0.736** | **0.589** | **0.679** | **0.604** | **0.691** |
| BERT$_{MFRC}$ | **0.663** | **0.533** | 0.312 | **0.396** | 0.272 | **0.491** | 0.619 | 0.475 | 0.296 | 0.321 | 0.250 | 0.392 |
| BERT$_{E2Mo}$ | 0.481 | 0.495 | **0.347** | **0.396** | **0.314** | 0.406 | 0.544 | 0.581 | 0.432 | 0.380 | 0.314 | 0.450 |

Table 2: F1-score performance results on the morality classification task. Columns correspond to test datasets.

| | SemEval | | | | | | | DailyDialogue | | | | | |
|---|---|---|---|---|---|---|---|---|---|---|---|---|---|
| | *anger* | *disgust* | *fear* | *joy* | *optimism* | *sadness* | overall F1 | *anger* | *disgust* | *fear* | *joy* | *sadness* | overall F1 |
| BERT$_{SemEval}$ | 0.579 | **0.557** | **0.399** | **0.607** | **0.461** | **0.418** | **0.51** | **0.448** | **0.153** | **0.147** | 0.689 | **0.466** | **0.559** |
| BERT$_{E2Mo}$ | **0.599** | **0.577** | 0.158 | 0.538 | 0.011 | 0.149 | 0.443 | 0.379 | 0.112 | 0.061 | **0.79** | 0.376 | 0.515 |

| | GoEmotion | | | | | | AffectiveText | | | | | |
|---|---|---|---|---|---|---|---|---|---|---|---|---|
| | *anger* | *disgust* | *joy* | *optimism* | *sadness* | overall F1 | *anger* | *disgust* | *fear* | *joy* | *sadness* | overall F1 |
| BERT$_{SemEval}$ | 0.196 | 0.138 | 0.176 | **0.169** | **0.128** | 0.161 | **0.205** | **0.114** | **0.331** | 0.477 | **0.316** | **0.268** |
| BERT$_{E2Mo}$ | **0.346** | **0.222** | **0.204** | 0.022 | 0.053 | **0.224** | 0.179 | 0.099 | 0.112 | **0.548** | 0.11 | 0.199 |

Table 3: F1-score performance results on the emotion classification task. Columns correspond to test datasets.

Table 3 presents the results on the aforementioned datasets. Similar to Table 2, the average F1 scores for each label and an overall averaged micro F1 score are shown. On SemEval, BERT$_{E2Mo}$ proves to be competitive with its in-domain counterpart for the labels *anger*, *anticipation*, *disgust*, and *joy*, while it reports very low scores for the labels *fear*, *optimism*, and *sadness*. This is not surprising, since *fear*, *optimism*, and *sadness* are underrepresented in E2MoCase, as shown in Figure 2a. This may have impacted on the training process of BERT$_{E2Mo}$, making it less effective to recognize examples with these emotions. Similarly, low performance on these labels are observed on the DailyDialogue, GoEmotions, and AffectiveText datasets. For the remaining labels, BERT$_{E2Mo}$ proves to be competitive, and in some cases superior, to BERT$_{SemEval}$ on these datasets. On DailyDialogue, BERT$_{E2Mo}$ recognizes the *joy* label, with an average F1 score of 0.79, and achieves an overall micro F1 score close to that of BERT$_{SemEval}$. On GoEmotion, BERT$_{E2Mo}$ outperforms BERT$_{SemEval}$ on the *anger*, *disgust*, and *joy* labels, and achieves the highest overall micro F1 score. On AffectiveText, BERT$_{SemEval}$ generally proves to be the better model. This could be due to the fact that this dataset is part of the SemEval 2007 competition, while BERT$_{SemEval}$ is trained on a dataset from the SemEval 2018 competition. Since they belong to the same competition, they may share common features. Nevertheless, BERT$_{E2Mo}$ achieves better performance than BERT$_{SemEval}$ on the *joy* label.

**Validation of events.** Finally, we provide preliminary tests on the validity of the events provided by E2MoCase. Unlike morality and emo-tions, events are not numerical scores but rather strings in JSON format containing actions and subjects involved. In this case, rather than evaluating a model trained on E2MoCase for event extraction—a complex task that would require a considerable and unnecessary implementation effort at this stage—we assess whether the actions, which serve as the *trigger* elements of an event, are useful for identifying emotions and moralities. More specifically, we previously lemmatize paragraphs and triggers of E2MoCase and encode them using TF-IDF vectorial representation. Hence, we train an MLP model for moral and emotion classification using paragraphs, triggers, or the concatenation of both as input. We choose TF-IDF as the encoding strategy instead of language models like BERT because the latter benefit from contextual information, which is absent in our representation of events. Table 4 shows the performance, in terms of F1 scores, obtained by using TF.IDF representation of paragraphs only (tf.idf+mlp$_{trig}$), of triggers only (tf.idf+mlp$_{trig}$), or both (tf.idf+mlp$_{par-trig}$) as input. It stands out that using in combination triggers and paragraphs improves performance in most cases as concerns the emotion classification task, and in all cases as concerns the moral value classification task.

| | Emotion Classification | | | | | | | | | Moral Classification | | | | | |
|---|---|---|---|---|---|---|---|---|---|---|---|---|---|---|---|
| | *anticipation* | *disgust* | *joy* | *optimism* | *anger* | *sadness* | *pessimism* | *fear* | overall F1 | *care* | *fairness* | *loyalty* | *authority* | *purity* | overall F1 |
| tf.idf+mlp$_{par}$ | <u>0.168</u> | <u>0.691</u> | <u>0.466</u> | **0.155** | <u>0.888</u> | <u>0.21</u> | <u>0.103</u> | <u>0.234</u> | <u>0.729</u> | <u>0.567</u> | <u>0.691</u> | <u>0.623</u> | <u>0.632</u> | <u>0.649</u> | <u>0.638</u> |
| tf.idf+mlp$_{trig}$ | 0.087 | **0.703** | 0.359 | 0.055 | 0.884 | 0.08 | 0.057 | 0.145 | **0.736** | 0.335 | 0.369 | 0.382 | 0.356 | 0.379 | 0.366 |
| tf.idf+mlp$_{par-trig}$ | **0.195** | 0.674 | **0.493** | <u>0.145</u> | **0.895** | **0.244** | **0.105** | **0.24** | 0.728 | **0.587** | **0.715** | **0.656** | **0.662** | **0.668** | **0.668** |

Table 4: F1-score performance results on the morality classification and emotion classification tasks, using E2MoCase as test dataset. trigger words extracted from events.

## 6 Conclusion

In this study, we proposed E2MoCase, the first dataset designed to link moral values, emotional responses and events to press narratives of prominent legal cases. These three aspects are pivotal for understanding how media biases shape public perception of social and legal issues. While the dataset was labeled using automatic procedures based on well-recognized domain-specific LLMs, our validation methodology showed that E2MoCase enables NLP tools to achieve competitive performance on human annotated datasets.

Future work will focus on expanding the dataset to include a wider array of sources and languages, increasing the robustness of the analysis. In particular, it would be interesting to link the retrieved news with official legal documents of the associated legal trials. We also aim to refine the tools used to detect events and classify emotional and moral content, thereby improving the precision with which biases can be identified.

## 7 Limitations

Besides recognizing the need for a more robust validation process support on human annotators, as well as the opportunity for maintaining E2MoCase with fresh news from SwissDox or other similar reliable sources, one further limitation is the linguistic scope. Currently, the collection only contains articles translated into English, which may result in the loss of complex meanings found in the original language versions. This may affect annotations of emotions and moral qualities, particularly in languages where cultural factors heavily influence emotional expression and moral interpretation. We emphasize that the original paragraphs of E2MoCase cover different languages, such as French, German and Italian.

Also, cross-cultural analysis of moral and emotional representations is another area that requires further exploration. The same legal case may evoke different moral and emotional responses in different countries due to cultural and social differences. Cross-cultural studies, such as retrieving news of a case from different countries' media, could provide insights into these differences, increasing the dataset's applicability and impact in global research contexts and providing a broader spectrum of media representations of legal narratives.

## Acknowledgements

C.M.G. and L.Z. were visiting the University of Lausanne mostly during the development of this work. The authors wish to thank Kyriaki Kalimeri for providing us with the MoralBERT models and training data.

# A Prompt example for case selection

We found the legal cases through both an extensive manual searching and the use of generative LLM tools with online search capabilities, particularly ChatGPT Plus, Perplexity, and Claude-3-Opus-200K. Below, we report an example of the prompts used on each tool:

*Search for newspaper articles related to judicial cases where judges or the media have been influenced by cultural or discriminatory stereotypes. The articles should focus on specific cases, such as those involving individuals who have experienced discrimination. For each case found, provide the name of the individual involved, the year it occurred, a description of the case, and the link to the source.*

# B SwissDox Query

```
query:
  dates:
   - from: 2020-01-01
   - to: 2024-03-16
   content:
       AND:
         - Johnny Depp
         - The Sun
result:
    format: TSV
    columns:
     - id
     - pubtime
     - medium_code
     - language
     - .........
```

Figure 5: SwissDox query example.

Figure 5 shows an example of query in YAML format regarding the legal case of Johnny Depp and The Sun. This query requires the retrieval of Swiss newspaper news in a specific time range (`2020-01-01` to `2024-03-16`, i.e. the current date at the time of the query). The result of the request is a TSV file containing the content of the article, in xml format, and other information such as document id (`id`), publication date (`pubtime`), media code (`medium_code`) and article language (`language`).

# C Selected cases

We selected 39 cases (Table 5) which received high coverage in Switzerland and was covered by at least one Swiss newspaper. Note that we prioritized recent cases, due to the significant developments in the legal system and shifts in public opinion regarding civil rights, discrimination, and biases,

| | | |
|---|---|---|
| Britney Spears Conservatorship | Charlie Gard | Troy Davis |
| Bill Cosby women accusation | Dominique Strauss-Kahn | Ahmaud Arbery |
| Johnny Depp and Amber Heard | Nils Fiechter and Adrian Spahr | Sandra Bland |
| Mike Ben Peter | Michael Brown | Tamir Rice |
| Harvey Weinstein | Philando Castile | Caster Semenya |
| Edward Snowden | Eric Garner | Eluana Englaro |
| Breonna Taylor | Ethan Couch | Mohamed Wa Baile |
| Chelsea Manning | Terri Schiavo | Kyle Rittenhouse |
| Rosemarie Aquilina | Julian Assange political | Starbucks Arrests |
| Trayvon Martin | O.J. Simpson | Robert Kelly |
| Amanda Knox | Jussie Smollett | Patrick Lyoya |
| Sarah Everard | Johnny Depp and The Sun | Cyntoia Brown-Long |
| Youssef Nada | Ahmed Mohamed | Dieudonné M'bala M'bala |

Table 5: Subjects of the selected legal cases.

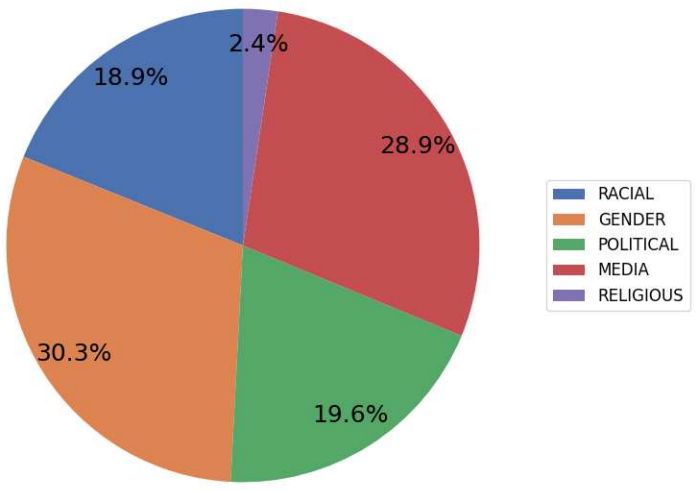

Figure 6: Distribution of the biases associated with the paragraphs. The bias of a paragraph is inherited from the legal case it originates from.

over the years. The biases in the legal cases have been manually detected and categorized as follows:

1. **Religious**, which refers to the unfair treatment or attitudes towards individuals or groups based on their religious beliefs.
2. **Media**, i.e., the tendency for media to present information in a way that reflects certain viewpoints, thereby influencing public perception.
3. **Political**, which involves favoring one political party, ideology, or candidate over others.
4. **Gender**, which refers to the preferential treatment or prejudice against individuals based on their gender.
5. **Racial**, which involves discriminatory attitudes or actions towards individuals based on their ethnicity.

The distribution of observed biases across these paragraphs is illustrated in Figure 6, with each paragraph inheriting the bias of the source news article.

# D Correlation Event

A Correlation Event is defined through the following Python class, according to the format required by GoLLIE:

```
class CorrelationEvent(Template):
  """A CorrelationEvent refers to any dynamics,
       state change or relationship within a context
        involving one or more entities. Such as
       discovering, accusing, loving, helping,
       killing,
  firing, finding, cheating, divorcing, reporting.
  """
  mention: str
  """
  The text span that most clearly expresses the
       CorrelationEvent. Such as: "discover", "kill
       ", "find", "love", "accuse", "love", "
       criticize", "discover", "fire", "cheat", "
       report"
  """
  entities: Dict[str, str] # Dictionary of
       CorrelationEvents in which the keys
       correspond to entities and values correspond
       to the role that entity has in the context of
        the CorrelationEvent. Such as: "Alex":"
       victim", "Claire": "police officer", "John":"
       man", "lawyer":"speaker"
```

Below are some examples of events extracted with GoLLIE based on the definition provided by the CorrelationEvent class:

```
{
  "mention": "trial",
  "entities": {
    "Weinstein": "subject"
  }
}

{
  "mention": "fire",
  "entities": {
    "Wilson": "Officer",
    "Michael Brown": "him"
  }
}

{
  "mention": "murder",
  "entities": {
    "Rittenhouse": "white supremacist"
  }
}
```

# E Outlier Removal and Cleaning process

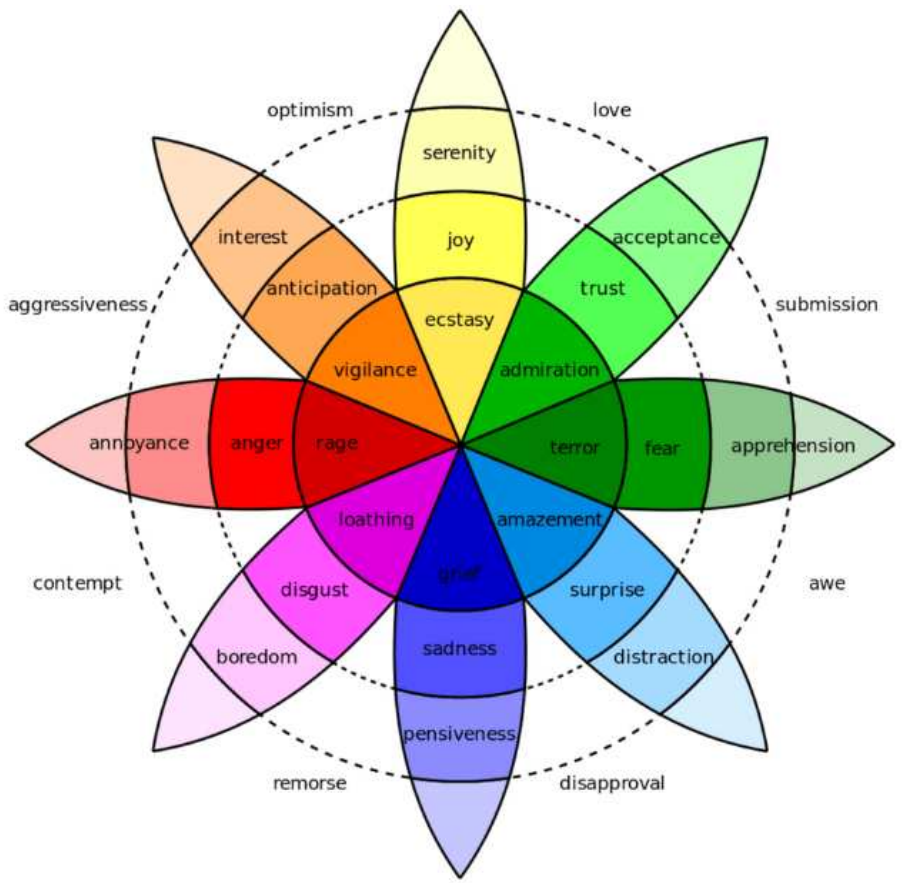

Figure 7: Plutchik's Wheel of Emotions. (Plutchik, 2001)

We follow the strategy described in Alg. 1 to detected and, eventually, remove possible outlier paragraphs. Specifically, given a specific case, we consider all paragraphs from news articles and the associated emotions. The most prevalent emotion $E$ is identified through all the paragraphs of the case. An emotion is prevalent if its scores boxplot has the largest inter quartile range (IQR). For example, considering Figure 8a, the prevalent emotion is *fear* as the IQR of its boxplot is the largest w.r.t. all the other boxplots.

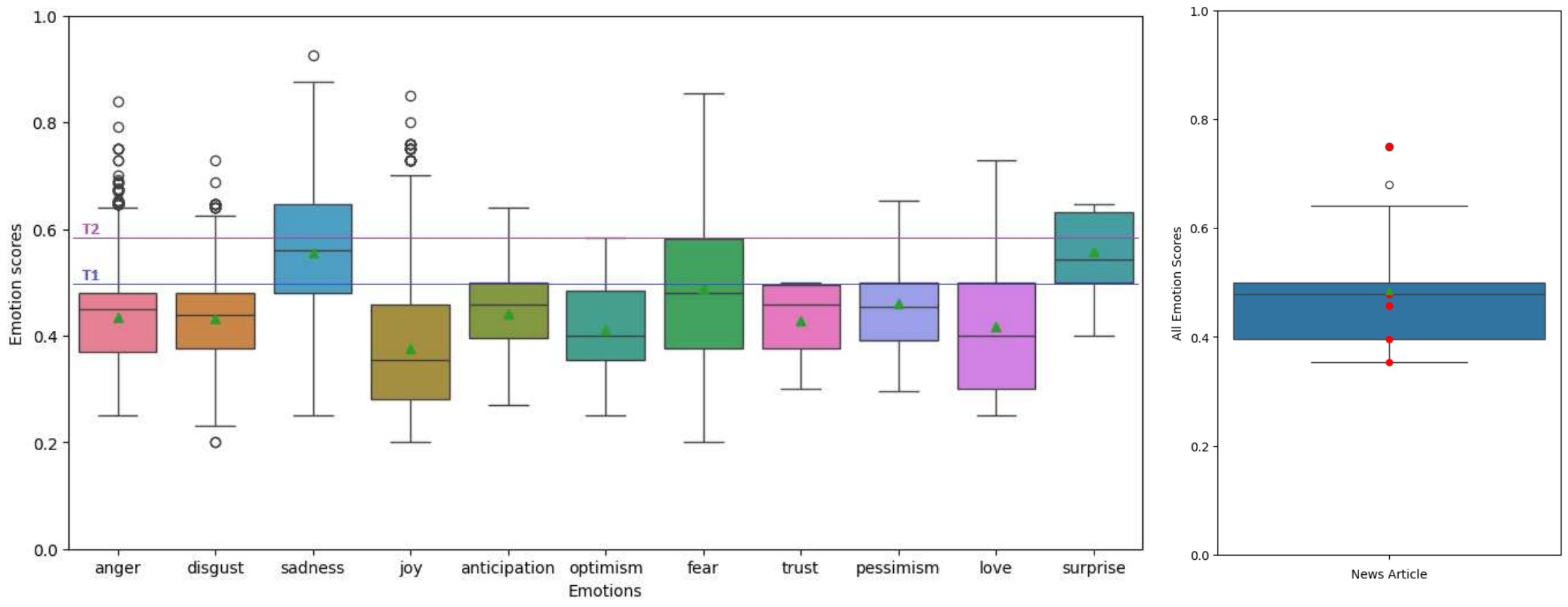

(a) (b)

Figure 8: (a) Boxplots of the emotions detected in the news article discussing Britney Spears conservatorship. The prevalent emotion is *fear*, from which $T1$ and $T2$ thresholds are retrieved. (b) Boxplot of all the emotions scores in the news article from which the candidate outlier paragraph comes from. The scores of the emotion associated to the candidate outlier are highlighted with red points.

Given $E$, we define its *opposite* and *contrary* emotions. An emotion *opposite* to $E$ is the emotion in the opposite position to $E$ in the Plutchik's wheel (Figure 7). For example, if $E$ is *joy*, its opposite emotion is *sadness*. If $E$ is a positive emotion (*joy*, *love*, *optimism*, *trust*), an emotion *contrary* to $E$ is an emotion with negative polarity (*anger*, *sadness*, *disgust*, *pessimism*, *fear*), and vice versa. Note that if an emotion is both opposite and contrary to $E$, it is considered solely as opposite. For example, if $E$ is *joy*, its contrary emotions are *anger*, *disgust*, *pessimism*, *fear* (*sadness* is already the opposite emotion).

Following Alg. 1, paragraphs with opposite or contrary emotions to $E$ are selected if their scores exceed a certain threshold, i.e. $T1$ for opposite emotions and $T2$ for contrary emotions. $T1$ is the mean value within the IQR of the prevalent emotion $E$, while $T2$ is the upper quartile of IQR. For example, in Figure 8a, we retrieve all the paragraphs with the opposite emotion *anger* whose scores exceed $T1$ and all the paragraphs with contrary emotion *joy*, *love*, *optimism* or *trust* whose scores exceed

```
Data: case_paragraphs,
top_emotion=get_emotion_with_max_IQR(case_paragraphs)
opposite_emotion=get_opposite_emotion(top_emotion)
contrary_emotions_list=get_contrary_emotions(top_emotion)
T1 = get_mean_from_max_IQR(case_paragraphs)
T2 = get_Q3_from_max_IQR(case_paragraphs)
Result: case with cleaned emotions
foreach case_paragraph in case_paragraphs do
    foreach emotion in case_paragraph do
        if emotion == opposite_emotion and
          score(emotion) > T1 then
            remove_if_outlier_in_news(case_paragraph,
            case_paragraphs,
            emotion)
        end
        if emotion in contrary_emotions_list and
          score(emotion) > T2 then
            remove_if_outlier_in_news(case_paragraph,
            case_paragraphs,
            emotion)
        end
    end
end
```

**Algorithm 1:** Candidate outlier paragraphs detection pseudocodice

$T2$. These paragraphs are considered candidates for being outliers, and they are analyzed following Alg. 2. Firstly, the entire article of the candidate outlier paragraphs $P$ is retrieved, and the emotions of all the article's paragraphs are evaluated. If the score of the opposite/contrary emotions in $P$ is not "common" compared to the scores of the entire article, it is considered as possibly anomalous. For example, Figure 8b show the boxplot of the scores of all the emotions of an article containing a candidate outlier paragraph, whose emotion

**Data:** case_paragraph,
case_paragraphs,
emotion,
T = 1.5,
news_article =
reconstruct_news_article(case_paragraph,
case_paragraphs)
Q1 = get_Q1_from_emotion_scores(news_article)
Q3 = get_Q3_from_emotion_scores(news_article)
**Result:** case_paragraphs (without outlier emotion)
**if** *score(emotion) < Q1 or score(emotion) > Q3* **then**
    lof_scores=LocalOutlierFactor(news_article)
    lower = mean(lof_scores) - T * std(lof_scores)
    upper = mean(lof_scores) + T * std(lof_scores)
    outliers = [index_of(s) for s in lof_scores if s < lower or s > upper]
    **if** *index_of(case_paragraph) in outliers* **then**
        remove(case_paragraph, case_paragraphs)
    **end**
**end**

**Algorithm 2:** Remove paragraph if outlier pseudocodice

(contrary or opposite) scores are highlighted as red points. If the emotion score of the candidate outlier paragraph is placed outside the IQR of the article boxplot, such paragraph is removed if it is anomalous according to the Local Outlier Factor (Breunig et al., 2000). The local outlier factor uses local density estimate with k nearest neighbors. It identifies regions with similar density and points with significantly lower density, identifying outliers by comparing the object's local density with its neighbors. In our case, we considered the embedding of each paragraph obtained with Sentence-Transformers (Reimers and Gurevych, 2019), specifically `all-mpnet-base-v1` since it has the best trade-off between maximum manageable text length, encoding speed and model size. The number of neighbors is dynamically calculated with respect to the total number of paragraphs in the current article, so as to ensure that the number of neighbors exceeds the minimum number of samples that a cluster must contain, but remains below the maximum number of neighboring samples that could potentially be local outliers. Following Alg. 2, a paragraph is removed if its emotion score exceeds the upper or lower bounds calculated starting from the mean value of the Local Outlier Factor scores. This approach ensures the robustness of our dataset by minimizing noise and enhancing the relevance of the analyzed content. This process resulted in the removal of about 100 paragraphs, which were manually checked to confirm that they were indeed outliers.

Additionally to the outlier removal process, we cleaned our labels by zeroing the values of morality and emotions with scores below a threshold (we empirically set it to 0.2). If, following this step, there were no more moral/emotional traits with values greater than zero, we assigned the corresponding paragraph the label of moral/emotional neutrality, with score 1. Considering the events, they passed through a lemmatization process, hence removing stop words, punctuation and digits. Then, we removed all repeated events within the same paragraph (i.e., same trigger word and same subjects involved) and all the paragraphs in which no event was detected.

## F Dataset analysis

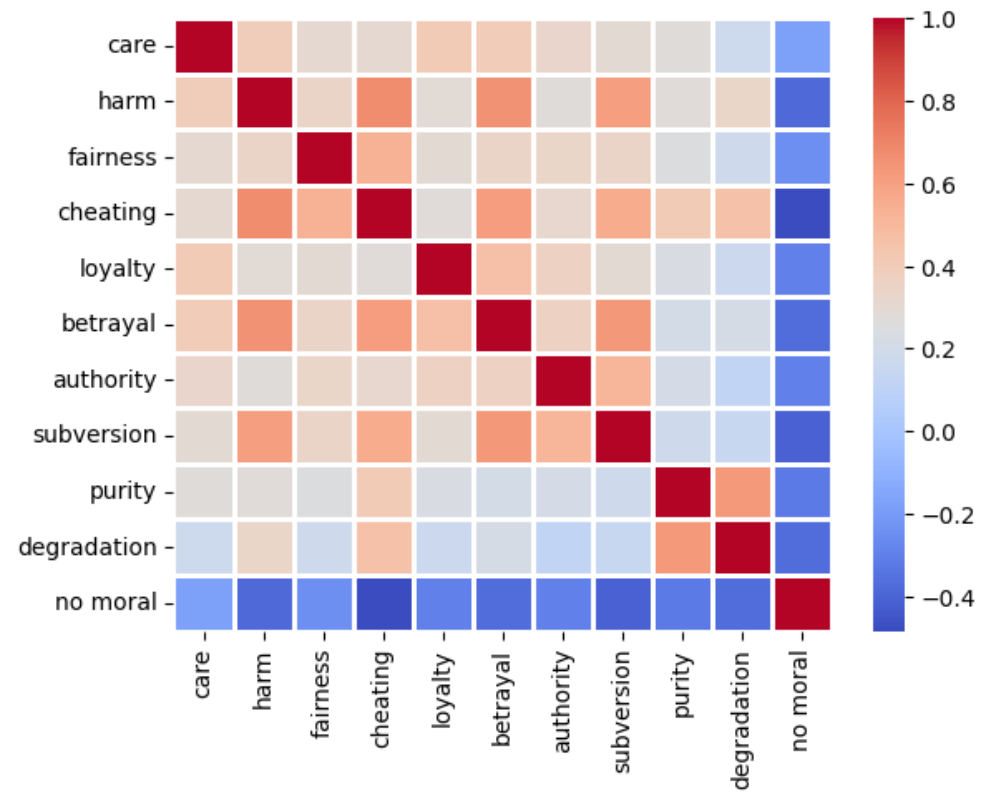

Figure 9: Correlation of morality labels in E2MoCase.

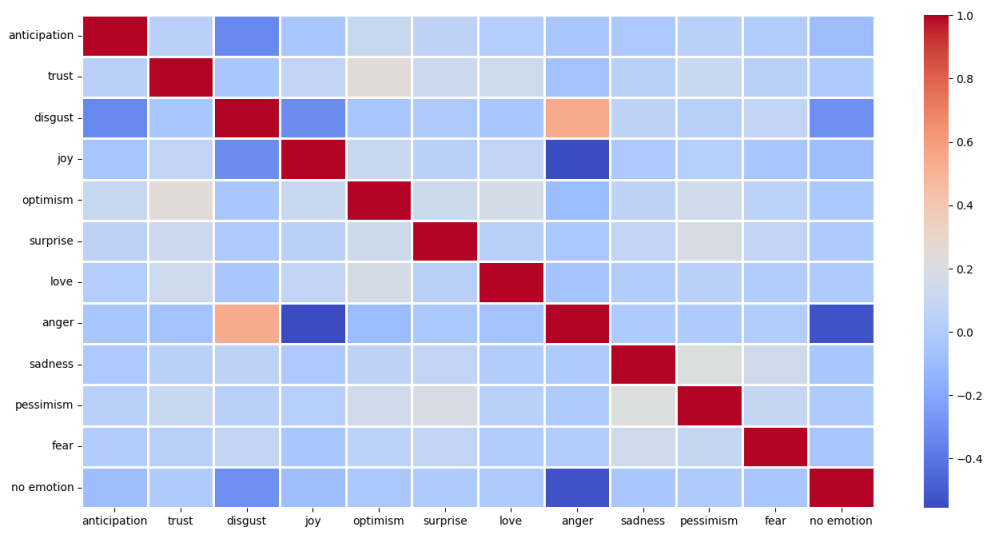

Figure 10: Correlation of emotion labels in E2MoCase.

Figure 9 and Figure 10 show, respectively, the correlation between morality and emotion labels in E2MoCase. As it can be noticed, 'negative' moralities (*harm*, *cheating*, *betrayal*, *subversion*) have a tendency to be correlated, while only a correlation between 'anger' and 'disgust' is evident for the emotions. This may be due to the different distribution of morality and emotions (Figures 2b and 2a). Furthermore, emotions tend to have a low score distribution, mostly below 0.5 (Figure 11).

Figure 12 shows the word cloud of paragraph events in which neither emotion nor morality were detected. Interestingly, in contrast to Figure 4 in which the events captured consisted of severe facts (e.g. murder, rape, trials), in this case the events treated mostly innocuous facts (e.g. buy, move, write) that potentially do not elicit strong emotions or express any particular moral values.

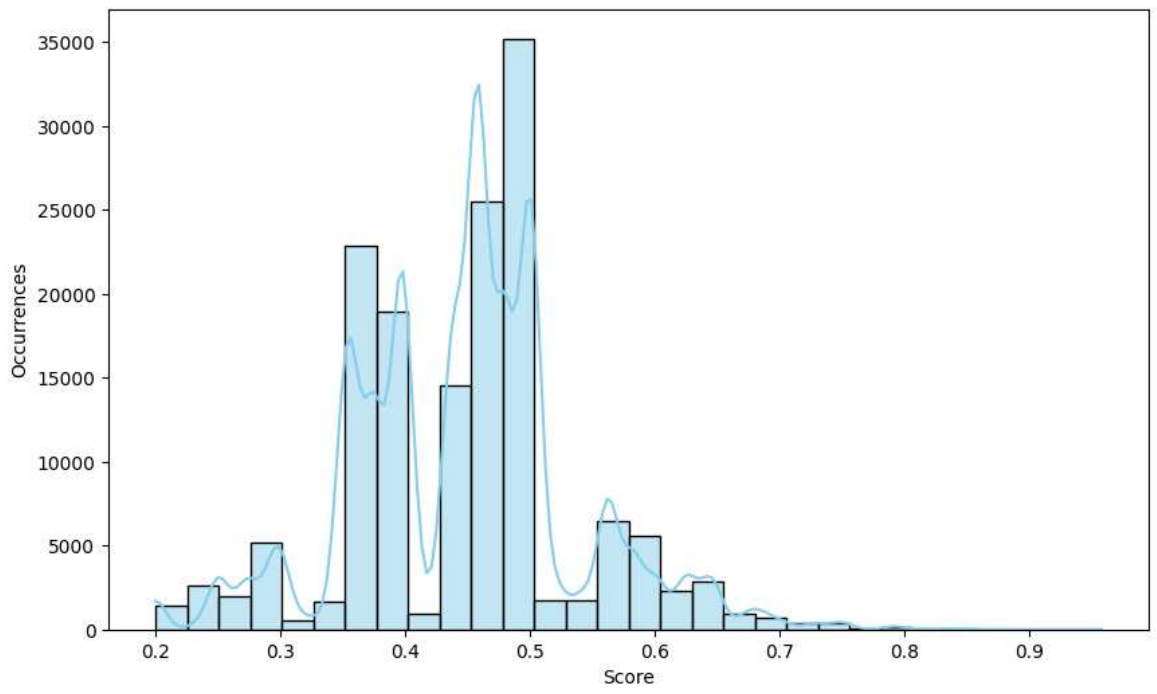

Figure 11: Distribution of emotion scores in E2MoCase.

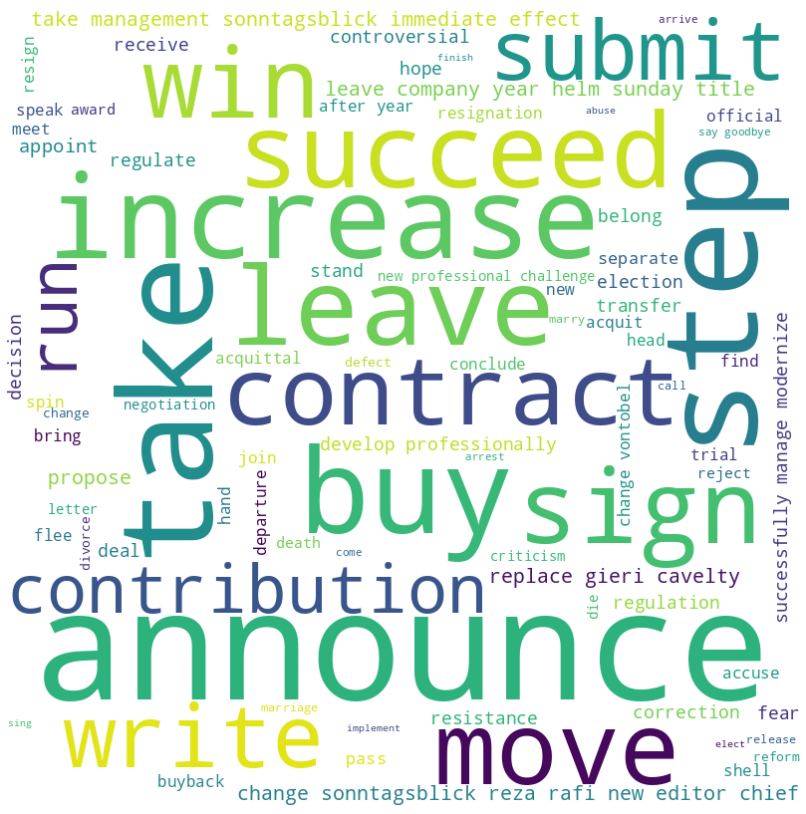

Figure 12: Word cloud of events for texts with no emotion and no morality.

Finally, Figure 13 shows similarity heatmaps for pairs of texts labeled with moral categories $(m_1, m_2)$ from the E2MoCase, MFTC, and MFRC datasets. In all three datasets, the labels are generated by MoralBERT. Note that MoralBERT is originally trained on MFTC and MFRC, hence the assigned labels for these datasets can be considered reasonably robust and, consequently, the corresponding heatmaps serve as a benchmark for comparison with the heatmap from E2MoCase. The three heatmaps show a high degree of similarity, suggesting firstly that MoralBERT consistently labels texts across the datasets, i.e. it assigns the same label $(m_1 = m_2)$ to texts with high similarity (likely discussing similar topics) and different labels $(m_1 \neq m_2)$ to texts with lower similarity.

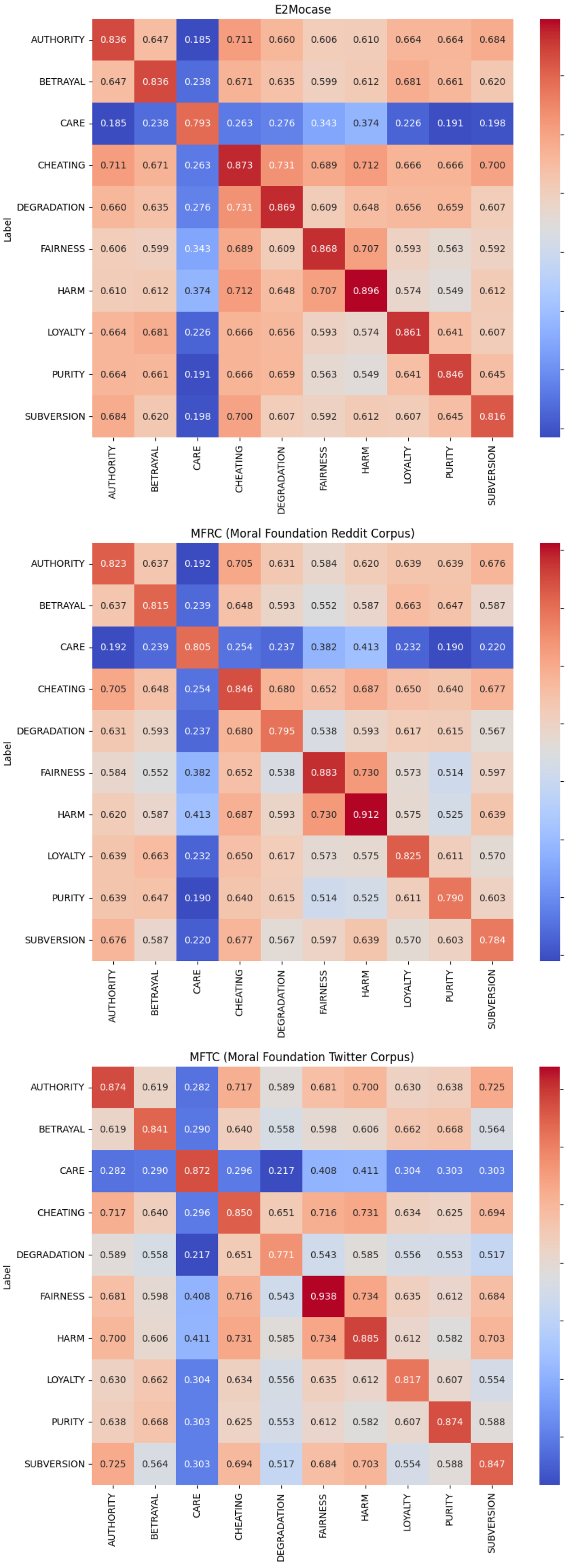

Figure 13: Similarity heatmaps of moralities on E2MoCase, MRC and MTFC